\pdfoutput=1

\documentclass[11pt]{article}

\usepackage{eacl2023}

\usepackage{times}
\usepackage{latexsym}

\usepackage[T1]{fontenc}

\usepackage[utf8]{inputenc}

\usepackage{microtype}

\usepackage{inconsolata}
\usepackage{booktabs}
\usepackage{pgfplots}
\pgfplotsset{compat=1.18}

\newcommand{\gain}[1]{\textcolor{teal}{#1}}
\newcommand{\loss}[1]{\textcolor{purple}{#1}}

\definecolor{custompurple}{RGB}{128, 0, 128}

\usepackage{graphicx}
\usepackage{xcolor}
\usepackage[most]{tcolorbox}

\definecolor{hitgreen}{RGB}{0,120,45}
\definecolor{ratered}{RGB}{190,0,0}
\definecolor{boxgray}{RGB}{247,247,249}
\definecolor{rulegray}{RGB}{180,180,186}

\newcommand{\hit}[1]{\textcolor{hitgreen}{\bfseries #1}}
\newcommand{\elide}{\textcolor{gray}{\,[\dots]\,}}
\newcommand{\rating}[1]{\textcolor{ratered}{\bfseries #1}}

\newtcolorbox{outbox}[1]{%
  breakable, enhanced, sharp corners=downhill,
  colback=boxgray, colframe=rulegray, boxrule=0.4pt,
  left=6pt, right=6pt, top=5pt, bottom=5pt,
  fonttitle=\small\bfseries, coltitle=black,
  colbacktitle=boxgray, title={#1},
  attach boxed title to top left={yshift=-2mm, xshift=4mm},
  boxed title style={colframe=rulegray, boxrule=0.4pt, sharp corners}
}

\definecolor{missred}{RGB}{178,24,43}

\newcommand{\miss}[1]{\textcolor{missred}{\bfseries #1}}

\usepackage{listings}
\usepackage[most]{tcolorbox}

\lstdefinestyle{promptstyle}{
  basicstyle=\ttfamily\scriptsize,
  breaklines=true,
  breakatwhitespace=true,
  breakindent=0pt,
  columns=fullflexible,
  keepspaces=true,
  showstringspaces=false,
  frame=none,
  aboveskip=0pt,
  belowskip=0pt,
  postbreak=\mbox{\textcolor{gray}{$\hookrightarrow$}\space},
}

\newtcolorbox{promptbox}[1][]{
  breakable, enhanced, sharp corners=downhill,
  colback=gray!3, colframe=black!65, boxrule=0.5pt,
  left=4pt, right=4pt, top=4pt, bottom=4pt,
  fonttitle=\bfseries\small, colbacktitle=gray!18, coltitle=black,
  attach boxed title to top left={yshift=-2pt, xshift=6pt},
  #1
}

\newenvironment{prompt}
  {\par\small\raggedright
   \setlength{\parskip}{4pt}\setlength{\parindent}{0pt}}
  {\par}

   \usepackage{amsmath,amssymb}
   \usepackage{algorithm}
   \usepackage{algpseudocode}

\title{Search-Tree-Grounded Explanations for Chess}

\author{
  Ahmed Nesar Tahsin Choudhury\textsuperscript{1},
  Abdul Mohaimen Al Radi\textsuperscript{2} \\
  {\bf Yu Tian\textsuperscript{2}, Md. Mosaddek Khan\textsuperscript{1}} \\
  \textsuperscript{1}Computer Science and Engineering, University of Dhaka \\
  \textsuperscript{2}Institute of Artificial Intelligence, University of Central Florida \\
  {\small\texttt{ahmednesartahsin-2020115612@cs.du.ac.bd}, \texttt{ab575577@ucf.edu},} \\
  {\small\texttt{yu.tian2@ucf.edu}, \texttt{mosaddek@du.ac.bd}}
}

\begin{document}
\maketitle
\begin{abstract}
Chess engines reliably identify strong moves but provide limited natural-language justification, whereas large language models (LLMs) communicate fluently but remain unreliable at chess search. We present TRACE-Chess, a training-free framework that separates search-based evidence discovery from explanation generation. Given a chess position, a chess engine constructs a search tree, which is augmented with human-plausible alternatives and reduced to a compact subgraph containing supporting and contrastive evidence. An LLM then transforms this evidence into an explanation through either recursive bottom-up aggregation or a lower-cost single-call approximation. We evaluate TRACE-Chess on a stratified benchmark of 100 Lichess puzzles using human assessment and LLM-as-a-judge evaluation. Relative to a position-only baseline, recursive generation improves aggregate human and LLM-as-a-judge scores by \textbf{6.6\%} and \textbf{12\%}, respectively. The single-call strategy achieves smaller gains but reduces token consumption by approximately \textbf{85\%}. These results show that structured engine search evidence can improve the accuracy and overall quality of LLM-generated chess explanations without additional training.

\end{abstract}

\section{Introduction}
\label{sec:intro}

\begin{figure}[t]
    \centering
    \includegraphics[width=\linewidth]{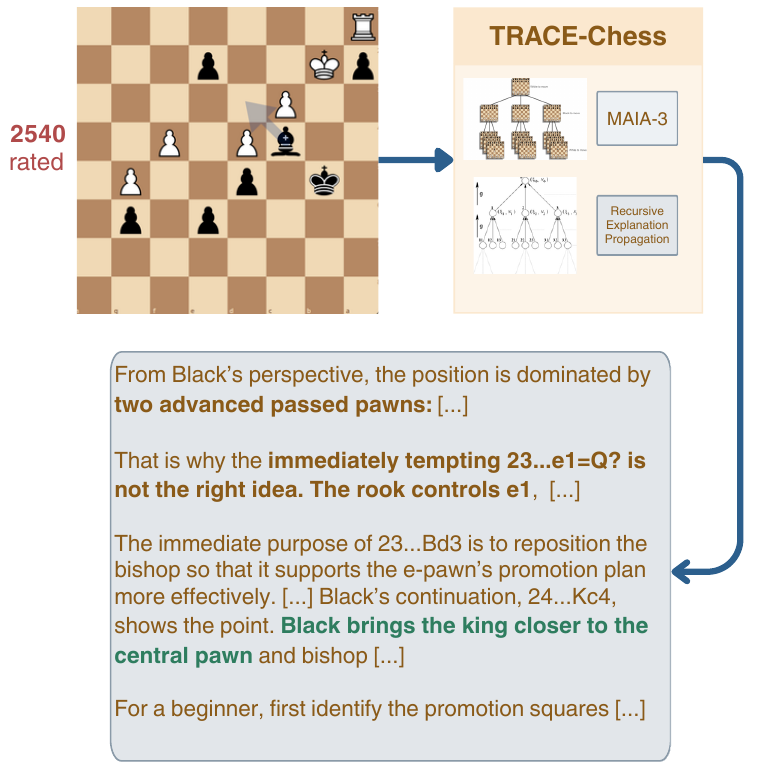}
    \caption{An example of search-tree-grounded chess explanation generation.}
    \label{fig:system_high_level}
\end{figure}

Chess has long served as a canonical testbed for artificial intelligence~\citep{Shannon1950,campbell2002deep}. Modern systems such as AlphaZero~\citep{silver2017masteringchessshogiselfplay} combine learned representations with Monte Carlo Tree Search (MCTS)~\citep{kocsis2006bandit} to discover strong strategies without handcrafted chess knowledge. Large language models (LLMs) introduce a complementary capability: communicating analyses and decisions in natural language. Recent work has therefore explored their use for chess move prediction, position evaluation, and commentary generation~\citep{monroe2024masteringchesstransformermodel,feng2023chessgptbridgingpolicylearning,tang2026groundedchessreasoninglanguage,kim-etal-2025-bridging}.

Most existing work focuses on selecting the best move from a board state, using either chess engines~\citep{silver2017masteringchessshogiselfplay,stockfish} or language models~\citep{monroe2024masteringchesstransformermodel,zhang-etal-2025-complete}. A strong move, however, may appear unintuitive to a human player, particularly when its value depends on a deeper continuation or the failure of a seemingly natural alternative. Chess commentary systems move closer to addressing this need, but typically produce brief descriptions of individual moves rather than complete justifications grounded in the alternatives considered during search~\citep{kim-etal-2025-bridging,jhamtani-etal-2018-learning}. Consequently, they provide limited support for explaining not only why the selected move succeeds, but also why plausible alternatives are inferior. This motivates our central research question: \textit{Can LLMs generate coherent chess explanations by grounding their analysis in evidence from an engine's search process instead of directly predicting the best move?}

We posit that \textit{chess explanation generation} should be viewed as a two-stage problem: (1) searching for evidence and (2) communicating evidence. These are fundamentally different capabilities. Chess engines are optimized for exploring the large space of possible continuations and evaluating candidate moves, whereas LLMs are well suited to interpreting structured information and expressing it in accessible natural language. Our training-free approach combines these complementary strengths: an engine supplies evidence from its search process, and an LLM transforms that evidence into a coherent, human-oriented explanation, as shown in Figure~\ref{fig:system_high_level}.

Given a chess position, our framework constructs a search tree using a chess engine and augments it with human-plausible moves. It then extracts relevant evidence for the engine-selected move and its alternatives. We explore two explanation strategies: recursive generation, where local explanations grounded in chess concepts (e.g. \textit{fork}, \textit{checkmate}, \textit{king safety}, etc.) are aggregated bottom-up into a strategic narrative, and a single-call variant that processes the selected search evidence in one prompt. The resulting explanation identifies the strongest move, explains why it works, and compares it with plausible alternatives.

We primarily evaluate the framework on 100 stratified Lichess puzzles~\citep{lichess} against a position-only baseline that explains the engine-selected move without search evidence. Human, LLM-based evaluations, and claim-level verifications show that search grounding improves aggregate explanation quality, particularly chess correctness and strategic-theme identification. Recursive generation generally performs best, while the single-call strategy is substantially more token-efficient.

We summarize our contributions as follows.

\begin{itemize}
    \item In contrast to prior work on \textbf{brief} chess commentary, we introduce a task requiring detailed, contrastive, and comprehensive explanations.

    \item We introduce a novel search-tree-grounded framework for chess explanation generation. Our training-free, recursive approach decouples chess search from explanation generation and grounds LLM-based explanations in structured evidence produced by a chess engine.

   \item Through human evaluation and LLM-as-a-judge assessment, we show that search-grounded explanations outperform a vanilla LLM prompting baseline, yielding aggregate improvements of \textbf{6.60\%} and \textbf{12\%}, respectively, across the evaluated dimensions. These results demonstrate that structured access to engine-generated reasoning traces can improve explanation quality without additional model training.
\end{itemize}

\section{Related Work}

This section reviews prior research in three key areas relevant to our work: best move prediction, chess commentary generation and evaluation.

\paragraph{Best Move Prediction.} 
Early computer-chess systems relied on symbolic search and handcrafted evaluation functions. Engines such as \emph{Deep Blue}~\citep{campbell2002deep} and \emph{Crafty}~\citep{hyatt1997crafty} used alpha--beta search together with chess-specific heuristics to identify strong moves. Modern engines such as \emph{Stockfish}~\citep{stockfish} extend this paradigm through highly optimized search, pruning, and evaluation. Reinforcement-learning systems such as \emph{AlphaZero}~\citep{silver2017masteringchessshogiselfplay} and \emph{MuZero}~\citep{Schrittwieser_2020} combine learned policy and value representations with tree search, while \emph{Leela Chess Zero}~\citep{lc0} applies an AlphaZero-style approach to chess. 

More recent studies have explored language models for chess move prediction by training on game records or prompting pretrained models to generate moves directly~\citep{feng2023chessgptbridgingpolicylearning}. Complementary work models the moves that human players are likely to choose, often accounting for player strength or behavioral patterns~\citep{monroe2026chessformerunifiedarchitecturechess,zhang2025allie}. Collectively, these approaches support either objectively strong move selection or human move prediction, but \textbf{provide limited support for translating evaluated continuations into coherent, human-interpretable justifications.}

\paragraph{Chess Commentary Generation.} 

Chess commentary generation aims to describe game events and strategic ideas in natural language. \citet{jhamtani-etal-2018-learning} introduced a large-scale dataset aligning chess positions and move sequences with commentary. More recent work, such as \emph{Concept-guided Chess Commentary Generation}~\citep{kim-etal-2025-bridging}, integrates expert models with LLMs to produce concept-oriented commentaries. These methods generally focus on concise commentary for individual moves and are less suited to explaining why a selected move is preferable to plausible alternatives or how its value emerges from deeper continuations. Producing \textbf{complete and contrastive explanations} from structured search evidence therefore remains \textbf{underexplored}.

\paragraph{Evaluation.} 

Standard text-generation metrics such as BLEU~\citep{papineni-etal-2002-bleu}, ROUGE~\citep{lin-2004-rouge}, and METEOR~\citep{banerjee-lavie-2005-meteor} measure similarity between generated and reference texts. Such metrics are poorly suited to chess explanations, where multiple valid explanations may differ substantially in wording and lexical overlap does not guarantee strategic correctness or adequate justification~\citep{kim-etal-2025-bridging}. Recent chess-commentary work has therefore complemented reference-based metrics with human judgments and LLM-based evaluation of task-specific explanation qualities~\citep{kim-etal-2025-bridging}.

\section{Methodology}
\label{sec:methodology}

\begin{figure*}[t]
    \centering
    \includegraphics[width=\textwidth]{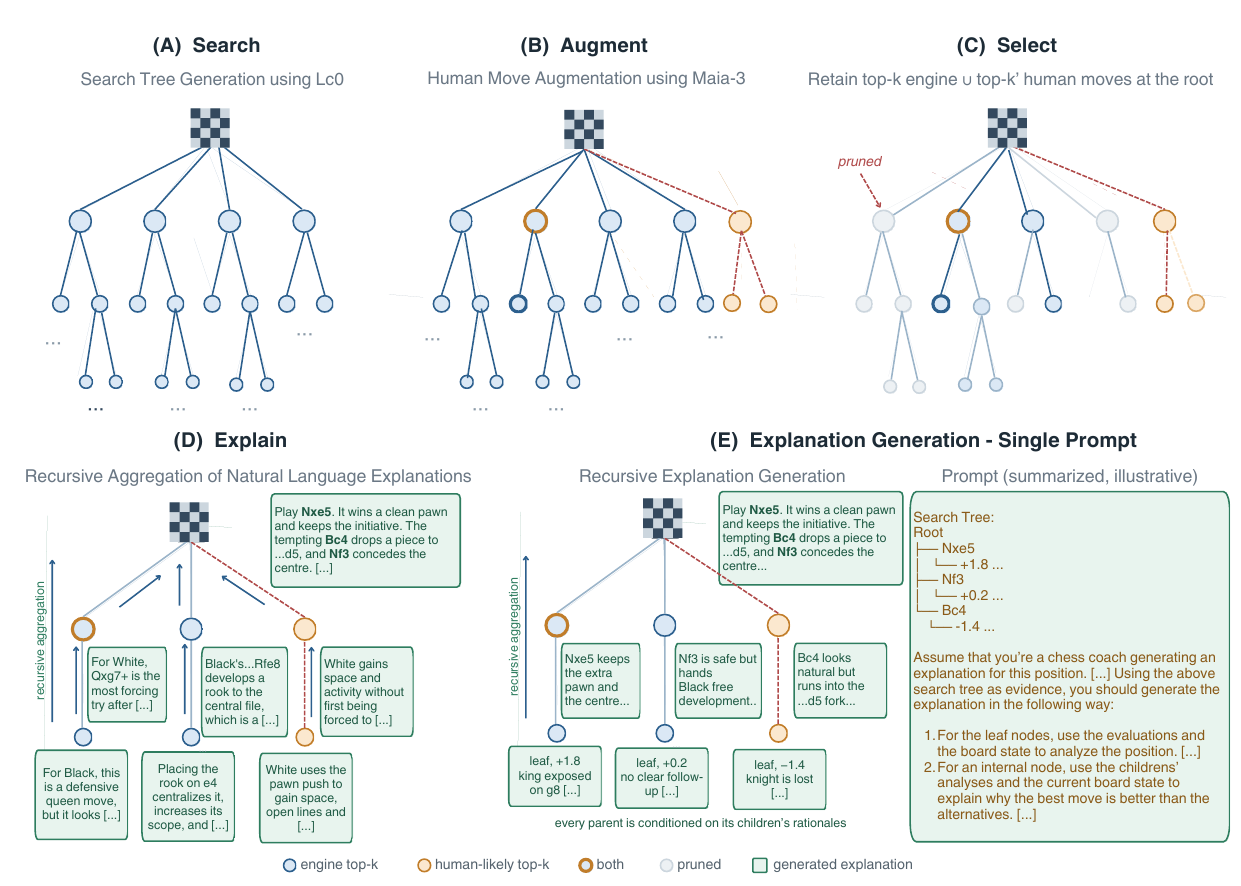}
    \caption{Overview of the proposed framework.
\textbf{(A) Search:} Lc0 expands an MCTS search tree from the root position.
\textbf{(B) Augment:} The search tree is augmented with human-plausible moves using Maia-3.
\textbf{(C) Select:} at the root we retain the union of the top-$k$ engine and top-$k'$ human-likely moves, yielding a compact subgraph.
\textbf{(D) Explain:} explanations are generated recursively over this subgraph: leaves from their evaluations and each internal node conditioned on its children's rationales, so the root explanation contrasts the best move against the alternatives.
\textbf{(E) Single prompt:} The serialized subgraph is explained in a single LLM call using the proposed algorithm.}
    \label{fig:pipeline}
\end{figure*}

\subsection{Problem Formulation}
\label{sec:problemformulation}

Given a chess position represented by a board state $s$ and the engine-selected best move $m^*$, our goal is to generate a natural language explanation $e$ that describes why $m^*$ is a strong choice. Unlike move prediction tasks, where the objective is to identify $m^*$ from $s$, we focus on the post-hoc explanation problem: $e = f(s, m^*, \mathcal{T})$, where $\mathcal{T}$ denotes structured search evidence generated by chess engines. A desirable explanation should be (1) \textit{strategically meaningful} to human players by correctly identifying the underlying strategic theme, (2) \textit{factually correct} with respect to chess rules and legal move sequences, and (3) \textit{comprehensive}, explaining not only why the selected move works but also why plausible alternative moves are inferior. 

\subsection{Search Tree Generation}

To generate structured search evidence, we use Leela Chess Zero (Lc0)~\citep{lc0}
as the primary search engine. Given an input position $s$, Lc0 uses
PUCT~\citep{rosin2011multi_puct} to produce a search tree $\mathcal{T}(s)$ containing candidate moves, resulting positions, evaluation estimates, and principal variations. Each node in $\mathcal{T}(s)$ represents a game state and edges correspond to legal moves connecting parent and child states. Figure~\ref{fig:pipeline}A illustrates an example of such a tree. A tree $\mathcal{T}(s)$ obtained in this way provides information about both the selected move and the alternatives considered by the engine. However, engines' preferred moves do not always align with moves that appear natural to human players. Therefore, to support \textit{contrastive} explanations, we first augment $\mathcal{T}(s)$ with \textit{human-plausible moves} before generating explanations.

\subsection{Search Tree Augmentation with Human Moves}

To incorporate human-oriented chess knowledge, we turn to Maia-3-79m~\citep{monroe2026chessformerunifiedarchitecturechess}, a neural chess model trained to predict human moves at different skill levels. Specifically, for each selected position, Maia-3 provides a set of candidate moves that represent realistic human considerations. 

Because some Maia-3 candidates may not appear in the original Lc0 search tree, we explicitly evaluate these moves using additional, light-weight Lc0 searches to obtain consistent search-based evidence for all candidate variations. Since the number of children of the root node can be large, we retain $K_e$ engine-provided candidate moves from Lc0 and $K_m$ human-plausible candidate moves from Maia-3 at the root position. Figure~\ref{fig:pipeline}B illustrates this procedure. Note that we use Maia-3 because it's recent and lightweight, but we could also use any other human-move prediction system, such as ALLIE~\citep{zhang2025allie}. Moreover, this approach also enables users to contrast the optimal move with any alternative move of their choosing.

\subsection{Search Evidence Subgraph Extraction}

The augmented search tree can contain thousands of nodes, which makes it impractical to provide the entire tree to an LLM. Therefore, we extract a compact subgraph containing the most relevant evidence for explanation generation.

We define two budgets: $B_{\text{best}}$ and $B_{\text{other}}$. $B_{\text{best}}$ controls the number of nodes selected along the principal variation of the engine-preferred move and $B_{\text{other}}$ controls the depth of expansion for alternative variations. We select nodes along the principal variation up to a depth $B_{\text{best}}$ and expand alternative branches from candidate moves up to depth $B_{\text{other}}$. The resulting graph (see Figure~\ref{fig:pipeline}C) contains both supporting evidence for the best move and contrastive evidence from alternative continuations.

\subsection{Recursive Explanation Generation}

Given the selected search subgraph $\mathcal{G}=(V,E)$, our goal is to generate a natural language explanation for the root position by recursively transforming search evidence into textual explanations. Each node $v \in V$ represents a chess position and contains the following information: $x_v = (s_v, a_v, \Delta_v, p_v)$, where $s_v$ is the board state at node $v$, $a_v$ is the move that transitions the parent node to $v$, $\Delta_v$ represents engine-provided evaluation information, and $p_v$ contains auxiliary information such as principal variations, move statistics, or human-plausible alternatives. For each non-leaf node $v$, let its children be denoted as: $C(v)=\{c_1,c_2,\dots,c_k\}$, where each child $c_i$ corresponds to a candidate continuation from position $s_v$. The explanation generation function produces an explanation $z_v$ for each node:
    $z_v = F(v)$, where $z_v$ is a textual description of the chess idea represented by node $v$. The final explanation for the root position $r$ is obtained as $z_r$.

\subsubsection{Leaf Node Explanation Generation}

Leaf nodes represent positions where no further explanation decomposition is performed. For a leaf node $l$, the LLM receives only the local chess information: $I_l = (x_l, m_l, \phi_l)$, where $m_l$ denotes the move leading to the leaf node and $\phi_l$ represents extracted chess features. Board features are extracted with rule-based heuristics implemented in the \texttt{python-chess} library, by comparing the positions before and after each candidate move. They cover material change, checks and captures, king safety (pawn shield, escape squares and attacked squares), piece development, central occupation, mobility (legal-move counts), pawn structure (passed, doubled and isolated pawns) and simple tactical motifs (forks and newly created pins). The leaf explanation is generated as: $z_l = \mathrm{LLM}(P_{\mathrm{leaf}}(I_l))$, where $P_{\mathrm{leaf}}$ is the leaf-level prompt template (see Appendix~\ref{app:prompts}).

\subsubsection{Internal Node Explanation Generation}

For an internal node $v$, the LLM receives both the local search information and explanations generated by its children. The input representation is:
    $I_v =
    \left(
    x_v,
    \{(x_{c_i}, z_{c_i})\}_{i=1}^{k}
    \right)$, where $z_{c_i}$ represents the explanation generated for child node $c_i$. The node explanation is then computed as:
    $z_v =
    \mathrm{LLM}(P_{\mathrm{internal}}(I_v))$ (see Appendix~\ref{app:prompts}).

The internal-node prompt instructs the LLM to synthesize the child explanations into a higher-level strategic narrative. Rather than independently analyzing every possible continuation, the model aggregates previously generated local insights and determines the central idea of the position (see Figure~\ref{fig:pipeline}D).

However, recursive generation requires one LLM call per node in the explanation subgraph. For a tree containing $|V|$ selected nodes, the total number of LLM calls is $|V|$, which can become expensive. To address this, we explore a single-call strategy in the following section that performs explanation synthesis over the entire selected subgraph in a single LLM invocation.

\subsubsection{Single-Call Explanation Generation}

To reduce the cost of recursive generation, we provide the entire extracted search evidence subgraph $\mathcal{G}=(V,E)$ to the LLM in a single prompt. The input is $I_{\mathrm{single}}=(x_r,\mathcal{G},\{x_v:v\in V\})$, where $x_r$ denotes the root position and $\mathcal{G}$ contains node states, move transitions, evaluations, and auxiliary search information. The final explanation is generated as $z_r=\mathrm{LLM}(P_{\mathrm{single}}(I_{\mathrm{single}}))$, where $P_{\mathrm{single}}$ includes both the search evidence and instructions for following the recursive explanation procedure: analyzing local information at leaf nodes, identifying tactical and strategic insights, and aggregating these insights toward the root position (see Appendix~\ref{app:prompts} to read the prompt). Figure~\ref{fig:pipeline}E illustrates the single-call strategy. Intuitively, the single-call formulation approximates recursive explanation generation by explicitly communicating the aggregation process within the prompt while reducing the number of LLM invocations to one.

\section{Experiments}

\subsection{Experimental Settings}

\paragraph{Model Selection.}
We select models using the LLM Chess Leaderboard\footnote{\url{https://maxim-saplin.github.io/llm_chess/}}, which ranks LLMs by their chess-playing competence \citep{kolasani2025llmchessbenchmarkingreasoning}. We initially explored using the top-ranked models, but they were impractical to run at the scale required for our experiments. We therefore restrict our selection to models that rank highly on the leaderboard while remaining tractable under our inference budget, yielding GPT-5.6 Luna and Grok-4.1-Fast-Reasoning for generating explanations and GPT-5.5 for judging, unless otherwise specified. We access these models through the Azure API. For a more detailed discussion on this, see Appendix~\ref{app:modelselection}.

\paragraph{Configuration Parameters.}
For each puzzle, we generate an Lc0 search tree containing 50K nodes. For human-move augmentation, we use the 79M-parameter variant of Maia-3 and set the target player rating to 1800. The default configuration retains $K_e=3$ engine-provided candidate moves and $K_m=2$ Maia-3 candidate moves from the root position, with search budgets $B_{\text{best}}=8$ and $B_{\text{other}}=4$. These values represent the standard configuration used throughout our experiments; any deviations from this setting are explicitly reported in the corresponding tables or results.

\begin{table*}[t]
\centering
\scriptsize
\setlength{\tabcolsep}{3pt}
\renewcommand{\arraystretch}{0.95}
\begin{tabular}{l ccccccc @{\hskip 6pt} ccc}
\toprule
& \multicolumn{7}{c}{\textbf{GPT-5.6 Luna}}
& \multicolumn{3}{c}{\textbf{Grok-4.1-Fast-Reasoning}} \\
\cmidrule(lr){2-8}\cmidrule(lr){9-11}
& \textbf{PO} & \textbf{PV} & \textbf{CCC}
& \multicolumn{2}{c}{\textbf{SC}}
& \multicolumn{2}{c}{\textbf{Rec}}
& \textbf{PO} & \multicolumn{2}{c}{\textbf{SC}} \\
\cmidrule(lr){2-2}\cmidrule(lr){3-3}\cmidrule(lr){4-4}
\cmidrule(lr){5-6}\cmidrule(lr){7-8}
\cmidrule(lr){9-9}\cmidrule(lr){10-11}
\textbf{Dimension}
& Score & Score & Score
& Score & $\Delta$\%
& Score & $\Delta$\%
& Score & Score & $\Delta$\% \\
\midrule
Chess Correctness
& 2.68 & 2.79 & 2.76
& \textbf{3.28} & \gain{+22.4}
& 2.89 & \gain{+7.8}
& 2.00 & 2.44 & \gain{+22.0} \\
Theme
& 3.51 & 3.91 & 3.59
& 4.08 & \gain{+16.2}
& \textbf{4.33} & \gain{+23.4}
& 3.36 & 3.77 & \gain{+12.2} \\
Pedagogy
& 3.73 & \textbf{3.94} & 3.80
& 3.48 & \loss{-6.7}
& 3.17 & \loss{-15.0}
& 2.90 & 2.32 & \loss{-20.0} \\
Completeness
& 3.04 & 3.25 & 3.05
& 3.45 & \gain{+13.5}
& \textbf{3.97} & \gain{+30.6}
& 2.88 & 3.24 & \gain{+12.5} \\
Contrastiveness
& 3.27 & 3.29 & 2.62
& 3.49 & \gain{+6.7}
& \textbf{3.86} & \gain{+18.0}
& 3.14 & 2.79 & \loss{-11.1} \\
\midrule
\textbf{Aggregate}
& 3.25 & 3.44 & 3.16
& 3.56 & \gain{+9.54}
& \textbf{3.64} & \gain{\textbf{+12.0}}
& 2.86 & \textbf{2.91} & \gain{\textbf{+1.75}} \\
\bottomrule
\end{tabular}
\caption{
Mean LLM-as-judge scores (1--5) for each evaluation dimension, comparing the two search-grounded strategies with three baselines. $\Delta$\% is relative to PO. Since the recursive strategy incurs substantially higher inference cost, and the PV/CCC baselines were run only with GPT-5.6 Luna, these are omitted for Grok. Best score per row for GPT-5.6 Luna is in bold. \textbf{Paired differences} and their \textbf{statistical significance} are reported in Table~\ref{tab:paired_diffs}.
}
\label{tab:model_strategy_dimensions}
\end{table*}

\begin{table*}[t]
\centering
\scriptsize
\setlength{\tabcolsep}{3pt}
\renewcommand{\arraystretch}{0.95}
\begin{tabular}{l ccc cc cc}
\toprule
& \multicolumn{3}{c}{\textbf{Mean Rating}}
& \multicolumn{2}{c}{\textbf{$\Delta$ vs.\ Baseline}}
& \multicolumn{2}{c}{\textbf{\% $\Delta$ vs.\ Baseline}} \\
\cmidrule(lr){2-4}
\cmidrule(lr){5-6}
\cmidrule(lr){7-8}
\textbf{Dimension}
& \textbf{PO} & \textbf{SC} & \textbf{Rec}
& \textbf{SC} & \textbf{Rec}
& \textbf{SC} & \textbf{Rec} \\
\midrule
Chess Correctness
& 3.56 & 3.96 & 4.12
& \gain{+0.40$^{*}$} & \gain{+0.56$^{*}$}
& \gain{+11.24\%} & \gain{+15.73\%} \\
Theme
& 3.96 & 4.24 & 4.40
& \gain{+0.28$^{*}$} & \gain{+0.44$^{*}$}
& \gain{+7.07\%} & \gain{+11.11\%} \\
Pedagogy
& 4.12 & 4.20 & 4.28
& \gain{+0.08} & \gain{+0.16$^{*}$}
& \gain{+1.94\%} & \gain{+3.88\%} \\
Completeness
& 4.35 & 4.42 & 4.54
& \gain{+0.07} & \gain{+0.19$^{*}$}
& \gain{+1.61\%} & \gain{+4.37\%} \\
Contrastiveness
& 4.40 & 4.44 & 4.48
& \gain{+0.04} & \gain{+0.08}
& \gain{+0.91\%} & \gain{+1.82\%} \\
\midrule
\textbf{Aggregate}
& 4.09 & 4.25 & \textbf{4.36}
& \gain{+0.16} & \gain{\textbf{+0.27}}
& \gain{+3.91\%} & \gain{\textbf{+6.60\%}} \\
\bottomrule
\end{tabular}
\caption{Average human evaluation ratings on a 1--5 scale. Improvements are measured relative to the position-only baseline. $^{*}$ indicates a statistically significant improvement over the PO baseline ($p<0.05$).}
\label{tab:human-evaluation}
\end{table*}

\providecommand{\ci}[1]{\textcolor{black!45}{#1}}          %
\providecommand{\pd}[3]{$#1$ \ci{[$#2$, $#3$]}}            %

\begin{table}[t]
\centering
\footnotesize
\setlength{\tabcolsep}{6pt}
\renewcommand{\arraystretch}{0.95}
\begin{tabular}{lc}
\toprule
\textbf{Strategy} & \textbf{Precision} \\
\midrule
Position Only & \pd{0.853}{0.826}{0.880} \\
Position+PV   & \pd{0.870}{0.846}{0.894} \\
Position+CCC  & \pd{0.878}{0.853}{0.903} \\
Single-Call   & \pd{0.887}{0.858}{0.916} \\
Recursive     & \pd{\mathbf{0.903}}{0.884}{0.922} \\
\bottomrule
\end{tabular}
\caption{Claim-level precision of explanations across the five strategies, with 95\% confidence intervals in gray. Precision is the fraction of extracted factual statements that pass deterministic verification. Best score is in bold.}
\label{tab:claim-precision}
\end{table}

\begin{table}[t]
\centering
\scriptsize
\setlength{\tabcolsep}{4pt}
\begin{minipage}[t]{0.42\columnwidth}
\centering
\begin{tabular}{lc}
\toprule
\textbf{Dimension} & \textbf{$\alpha$} \\
\midrule
Chess Correctness  & 0.55 \\
Theme              & 0.49 \\
Pedagogy           & 0.31 \\
Completeness       & 0.33 \\
Contrastiveness    & 0.28 \\
\bottomrule
\end{tabular}
\captionsetup{justification=raggedright,singlelinecheck=false}
\caption{Inter-rater agreement across the five evaluation dimensions.}
\label{tab:krippendroffs-alpha}
\end{minipage}
\hfill
\begin{minipage}[t]{0.55\columnwidth}
\centering
\begin{tabular}{lcc}
\toprule
\textbf{Dimension} & \textbf{$\rho$} & \textbf{$\tau$} \\
\midrule
Chess Correctness  & 0.46 & 0.31 \\
Theme              & 0.43 & 0.38 \\
Pedagogy           & 0.15 & 0.11 \\
Completeness       & 0.29 & 0.28 \\
Contrastiveness    & 0.25 & 0.17 \\
\bottomrule
\end{tabular}
\captionsetup{justification=raggedright,singlelinecheck=false}
\caption{Agreement between LLM-as-a-Judge and human annotations, measured by Pearson's $\rho$ and Kendall's $\tau$.}
\label{tab:judge_agreement}
\end{minipage}
\end{table}

\begin{figure}[t]
\centering
\begin{tikzpicture}
\begin{axis}[
    ybar,
    width=\columnwidth,
    height=3.6cm,
    bar width=5pt,
    ymin=-0.25,
    ymax=0.68,
    ylabel={Avg.\ gain},
    ylabel style={font=\scriptsize, yshift=-4pt},
    ytick={-0.2,0,0.2,0.4,0.6},
    yticklabel style={font=\scriptsize},
    symbolic x coords={
        501--1000,
        1001--1500,
        1501--2000,
        2001--2500,
        2501--3000
    },
    xtick=data,
    xticklabel style={font=\scriptsize},
    enlarge x limits=0.12,
    ymajorgrids=true,
    grid style={dashed, gray!50},
    axis x line*=bottom,
    axis y line*=left,
    extra y ticks={0},
    extra y tick labels={},
    extra y tick style={grid=major, major grid style={solid, black}},
    legend style={
        at={(0.02,0.98)},
        anchor=north west,
        font=\tiny,
        inner sep=1.5pt,
        draw=black!35,
        row sep=-2pt,
    },
    legend image post style={scale=0.6},
]
\addplot[fill=blue!60] coordinates {
    (501--1000,-0.17) (1001--1500,0.38) (1501--2000,0.48)
    (2001--2500,0.53) (2501--3000,0.33)
};
\addplot[fill=orange!70] coordinates {
    (501--1000,0.20) (1001--1500,0.41) (1501--2000,0.47)
    (2001--2500,0.52) (2501--3000,0.39)
};
\legend{Single-call, Recursive}
\end{axis}
\end{tikzpicture}
\vspace{-4pt}
\caption{Average gain in aggregate LLM-as-a-Judge score over the position-only baseline, for single-call and recursive strategies, across Lichess puzzle rating groups.}
\label{fig:rating_group_gain}
\vspace{-6pt}
\end{figure}
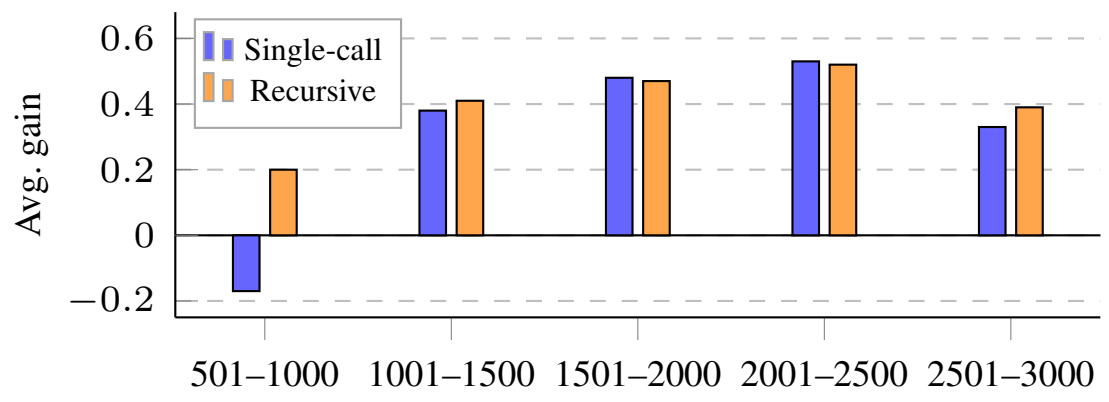

\begin{table*}[t]
\centering
\scriptsize
\setlength{\tabcolsep}{4pt}
\begin{tabular}{lccccccc}
\toprule
\textbf{Strategy} &
\shortstack{\textbf{Board-state}\\\textbf{errors}} &
\shortstack{\textbf{Illegal}\\\textbf{moves}} &
\shortstack{\textbf{Incorrect tactics}\\\textbf{or long-term reasoning}} &
\shortstack{\textbf{Incorrect contrastive}\\\textbf{move explanations}} &
\shortstack{\textbf{Implausible}\\\textbf{candidate moves}} &
\shortstack{\textbf{Unsupported}\\\textbf{causal claims}} &
\shortstack{\textbf{Positional}\\\textbf{misunderstanding}} \\
\midrule
Position Only & 20 & 16 & 28 & 20 & 16 & 24 & 24 \\
Single-Call   & 16 & 12 & 20 & 12 & 12 & 16 & 16 \\
Recursive     & \textbf{8} & \textbf{8} & \textbf{12} &
                \textbf{8} & \textbf{8} & \textbf{8} & \textbf{12} \\
\bottomrule
\end{tabular}
\caption{Error rates (\%) across the position-only baseline, the single-call strategy and the recursive strategy. Each value is the percentage of explanations containing at least one error of the indicated type. \textbf{Lower is better}.}
\label{tab:baseline_error_analysis}
\end{table*}

\subsection{Dataset Construction}

We construct our evaluation dataset from Lichess chess puzzles, each of which includes an initial board state, the side to move, popularity, rating, theme, and a solution sequence. Because evaluating frontier LLMs requires costly long-form generation across multiple models, we build a benchmark of \textbf{100 positions}. To assess performance across different levels of puzzle difficulty, we divide the puzzles into five Lichess rating intervals spanning 501 to 3000 and sample \textbf{20 puzzles from each interval}. Within each interval, puzzles are ranked by popularity before selection~\footnote{Random puzzle selection is discussed in Appendix~\ref{app:random-puzzle-selection}.}. This stratified design enables a controlled analysis of whether search-grounded explanation generation remains effective across difficulty levels.

\subsection{Baselines}

We primarily compare against the \textbf{Position Only} baseline: it receives the chess position, the side to move, extracted board features, and the engine-selected best move. Additionally, to test how much recursive summarization helps, we consider the following baselines: \textbf{1) Position + PV} receives the serialized principal variation, and \textbf{2) Position + Concepts} receives the chess concepts predicted by a classifier trained with the CCC codebase~\citep{kim-etal-2025-bridging}, as a point of comparison with CCC. Because expert annotation is costly (5--6 hours per annotator), our human study covers only the Position Only baseline, and we evaluate the other two with the LLM judge.

\subsection{Evaluation Metrics}

Evaluating chess explanations requires assessing qualities beyond lexical similarity, including correctness, strategic insight, pedagogy, completeness, and contrastiveness. We therefore use two complementary protocols: human evaluation for detailed assessment and LLM-as-a-judge evaluation for scalable evaluation across the full benchmark.

\paragraph{LLM-as-a-Judge Evaluation.}

While human evaluation provides the most reliable assessment of explanation quality, it is costly and difficult to scale. Traditional text-generation metrics (e.g. BLEU) are unsuitable for this task because multiple explanations may be correct despite using different wording. Following recent work on evaluating chess commentary with LLM-based judges~\citep{kim-etal-2025-bridging}, we use a task-specific rubric designed for chess explanations using LLM-as-a-Judge. The judge uses a \textbf{5-point Likert scale} across five dimensions: \textit{correctness}, \textit{theme}, \textit{pedagogy}, \textit{completeness}, and \textit{contrastiveness}. The complete prompt is provided in Appendix~\ref{app:prompts}. 

\paragraph{Human Evaluation.}
We conduct a human evaluation with four participants recruited from the university campus. All human annotators were recruited voluntarily and compensated at an hourly rate above the local minimum wage in their jurisdiction. All participants have a chess.com Rapid rating above \textbf{1450}, with an average rating of \textbf{1512}. Each participant evaluates explanations for the \textbf{same 25} positions, comprising \textbf{5} positions from each puzzle-rating bracket. For each position, participants assess outputs from three explanation strategies (position-only, single-call, and recursive), resulting in 75 assessments per participant and 300 assessments overall. The evaluation requires approximately \textbf{5--6 hours} per participant because the explanations are detailed and require careful examination. Participants compare the generated explanations and rate each one on a \textbf{5-point Likert scale} across the same five dimensions as the LLM-judge. Dimension-specific anchors are provided in Appendix~\ref{app:evaldimensions}. The complete evaluation protocol and participant instructions appear in Appendix~\ref{app:humanevalprotocol}.

\paragraph{Claim-Level Verification.} To address potential inaccuracies in LLM-based correctness judgments, we also perform claim-level verification. Inspired from a cotemporary work by \citet{hebbar2026hallucinationsboardtoolaugmentedevaluation}, we prompt an LLM (Appendix~\ref{app:claim-level-verification-prompt}) to extract claims from each explanation, decompose them into factual statements, and select an appropriate verification tool for each statement. These statements are then checked deterministically using \texttt{python-chess}. Note that this approach verifies concrete board-level facts (e.g., whether a rook attacks the queen) but does not necessarily establish the move's underlying purpose.  For example, a rook move may attack the queen while serving a deeper purpose (e.g., supporting a pawn push). An explanation that correctly identifies the attack but misses this underlying idea can still achieve high claim-level precision.

\subsection{Main Results}
\label{subsec:results}

\paragraph{LLM-as-a-Judge Evaluation.}

Table~\ref{tab:model_strategy_dimensions} compares the single-call and recursive strategies with the baselines under LLM-as-a-judge evaluation. With GPT-5.6 Luna, the recursive strategy achieves the highest overall score, followed closely by the single-call strategy. Across both backbone models, high gains appear in chess correctness, completeness and theme identification, suggesting that search-tree evidence most strongly improves the factual and thematic grounding of explanations. Pedagogy scores, however, decline for both models, while contrastiveness also decreases for Grok-4.1-Fast-Reasoning. Manual inspection suggests two possible reasons: grounded explanations use more technical chess terminology, for which the LLM judge assigns lower pedagogy scores; and, in the single-call setting, the model must cover a long sequence of evidence within one generation, which can lead to compressed explanations and omitted contrastive comparisons. \textit{What if the judge is provided the serialized search tree?} We find that similar trends persist (Appendix~\ref{app:judge-with-search-tree}).

\paragraph{Human Evaluation.}

Table~\ref{tab:human-evaluation} shows that both search-grounded generation strategies outperform the position-only baseline overall. The \textbf{recursive strategy} yields the strongest improvements, particularly in dimensions requiring deeper reasoning, such as strategic-theme identification and completeness. The single-call strategy also performs competitively, indicating that access to structured search evidence can improve explanation quality even without explicit node-by-node aggregation. Appendix~\ref{app:casestudy} presents a qualitative case study of a Lichess puzzle, comparing the explanations produced by the three strategies in detail and showing how the recursive strategy can uncover long-term motifs.

We measure \textbf{inter-rater agreement} using Krippendorff's $\alpha$~\citep{krippendorff1970estimating} with an ordinal distance metric, reflecting the ordered nature of the rating scale. As reported in Table~\ref{tab:krippendroffs-alpha}, agreement is higher for chess correctness and theme identification than for pedagogy, completeness, and contrastiveness. This suggests that correctness and thematic content are judged more consistently across raters, whereas the remaining dimensions involve more subjective assessments of explanation quality.

To assess \textbf{agreement between human and LLM ratings}, Table~\ref{tab:judge_agreement} reports Pearson's $\rho$ and Kendall's $\tau$. As with inter-rater agreement, the strongest correlations occur for \textit{chess correctness} and \textit{theme}, whereas the remaining dimensions show weaker alignment. Manual inspection suggests that the LLM judge follows the provided rubric more rigidly, while human evaluators apply the criteria more flexibly and are better able to account for their underlying intent.

\providecommand{\ci}[1]{\textcolor{black!45}{#1}}  %

\paragraph{Claim-Level Verification.}
Table~\ref{tab:claim-precision} reports claim-level precision across strategies. Precision is high for all strategies (above 0.85), suggesting that most factual statements in the generated explanations are board-accurate. The recursive strategy achieves the highest precision (0.903), a $+0.05$ improvement over position-only (0.853), and their 95\% confidence intervals do not overlap. Single-call (0.887) also improves over position-only, though the intervals overlap, as they do among the position-only, PV, CCC, and single-call strategies. This indicates search-grounded generation reduces factual errors at the claim level. We note that high precision does not imply that an explanation captures the move's underlying purpose. Our manual analysis indicates that recursive generation more frequently captures this underlying purpose, a distinction not measured by claim-level precision.

\subsection{Analysis}
\label{sec:analysis}

\paragraph{Maia-3 Ablation.}
At a matched candidate budget, incorporating Maia-3 improves \textbf{contrastiveness} by $+0.14$ $[+0.04, +0.24]$ ($p < 0.05$) for single-call and $+0.36$ $[+0.19, +0.53]$ ($p < 0.001$) for recursive. It also yields gains in \textbf{completeness} (Appendix~\ref{app:paired-differences}). This suggests that human-plausible moves from Maia-3 make explanations more contrastive and, to a lesser extent, more complete.

\paragraph{Grounding Gains Grow with Puzzle Difficulty.}
Figure~\ref{fig:rating_group_gain} reports the average improvement in aggregate LLM-as-a-judge score over the position-only baseline, disaggregated by Lichess puzzle rating.\footnote{We use LLM-as-a-judge scores for this analysis because the human evaluation includes five puzzles per rating band (25 in total), which is insufficient for reliable per-band comparisons.} The pattern suggests that grounding becomes more valuable as positions become less straightforward to interpret. For relatively simple puzzles, the central motif and refutation may already be apparent from the board, leaving less room for additional search evidence to help. As difficulty increases, explanations depend more heavily on deeper continuations that are not evident from the initial position, allowing the grounded strategies to outperform the position-only baseline by a wider margin.

\paragraph{Error Analysis of Generated Explanations.}Table~\ref{tab:baseline_error_analysis} illustrates a consistent reduction in errors from position-only to single-call and recursive generation using the human evaluation dataset. Search-tree-grounding reduces factual mistakes, implausible alternatives, and unsupported explanations, while recursive aggregation further improves tactical and positional reasoning. These trends demonstrate that structured search evidence and intermediate summarization can improve explanation reliability.

Additionally, we perform analysis of \textbf{human and LLM judge disagreements}, which appears in Appendix~\ref{app:human-llm-disagreement-analysis}. We also find that there is a \textbf{performance-cost trade-off} between the single-call strategy and the recursive strategy. The corresponding analysis is presented in Appendix~\ref{app:cost-analysis}.

\section{Conclusions}
\label{sec:conlusion}

We study \textit{chess explanation generation} by separating evidence discovery from communication.
Our \textit{training-free, search-tree-grounded} framework combines chess engines’ search capabilities with LLMs’ explanatory abilities.
Experiments show improved explanation quality, with generally larger gains on harder puzzles.
These findings suggest that structured search evidence may also support LLM explanations in domains such as theorem proving and strategic planning.

\section*{Limitations}

\paragraph{Single domain and single engine family.}
All of our experiments are conducted in chess, using search trees serialized
from Lc0 and human-plausibility scores from Maia-3. While the subgraph
extraction and recursive generation machinery is in principle
domain-agnostic, it assumes only a rooted search tree with per-node statistics, we do not empirically demonstrate transfer to other
MCTS domains (Go, Shogi, planning benchmarks) and leave it for future work.

\paragraph{Evaluation relies on LLM-as-a-Judge.}
Our primary evaluation uses an LLM judge with a five-dimensional rubric. While LLM-based judgments provide scalable evaluation, they may not perfectly align with expert human assessments. Although we additionally conduct human evaluation, scaling expert annotation remains challenging due to the substantial time and domain expertise required for assessing chess explanations.

\section*{Ethics Statement}

\paragraph{Human evaluation.}
Our human agreement study involved human annotators recruited on a
voluntary basis from the university campus. Annotators
were informed of the study's purpose, told that their ratings would be used
to evaluate a chess explanation generation system, and free to withdraw at any point. The task involved reading chess explanations and
carried no foreseeable risk of harm. No demographic or personal data was collected, and ratings are released in aggregate only.

\paragraph{Modeling human play.}
Our subgraph selection uses a human-move prediction model to prioritize lines
a person would plausibly consider. Such models are trained on the play of
particular populations and may encode the stylistic tendencies and blind
spots of those populations. Explanations conditioned on this signal
may consequently be better calibrated for some players than others. We do
not attempt to characterize this disparity here.

\bibliographystyle{acl_natbib}
\bibliography{custom}

\begin{thebibliography}{24}
\expandafter\ifx\csname natexlab\endcsname\relax\def\natexlab#1{#1}\fi

\bibitem[{Banerjee and Lavie(2005)}]{banerjee-lavie-2005-meteor}
Satanjeev Banerjee and Alon Lavie. 2005.
\newblock \href {https://aclanthology.org/W05-0909/} {{METEOR}: An automatic
  metric for {MT} evaluation with improved correlation with human judgments}.
\newblock In \emph{Proceedings of the {ACL} Workshop on Intrinsic and Extrinsic
  Evaluation Measures for Machine Translation and/or Summarization}, pages
  65--72, Ann Arbor, Michigan. Association for Computational Linguistics.

\bibitem[{Campbell et~al.(2002)Campbell, Hoane, and Hsu}]{campbell2002deep}
Murray Campbell, A.~Joseph Hoane, and Feng-hsiung Hsu. 2002.
\newblock Deep blue.
\newblock \emph{Artificial Intelligence}, 134(1--2):57--83.

\bibitem[{Feng et~al.(2023)Feng, Luo, Wang, Tang, Yang, Shao, Mguni, Du, and
  Wang}]{feng2023chessgptbridgingpolicylearning}
Xidong Feng, Yicheng Luo, Ziyan Wang, Hongrui Tang, Mengyue Yang, Kun Shao,
  David Mguni, Yali Du, and Jun Wang. 2023.
\newblock \href {http://arxiv.org/abs/2306.09200} {Chessgpt: Bridging policy
  learning and language modeling}.

\bibitem[{Hebbar et~al.(2026)Hebbar, Sheng, Oh, and
  Viswanath}]{hebbar2026hallucinationsboardtoolaugmentedevaluation}
S.~Ashwin Hebbar, Peiyao Sheng, Sewoong Oh, and Pramod Viswanath. 2026.
\newblock \href {http://arxiv.org/abs/2608.04240} {Hallucinations on the board:
  Tool-augmented evaluation of llm chess commentary}.

\bibitem[{Hyatt and Newborn(1997)}]{hyatt1997crafty}
Robert~M. Hyatt and Monroe Newborn. 1997.
\newblock Crafty goes deep.
\newblock \emph{ICCA Journal}, 20(2):79--86.

\bibitem[{Jhamtani et~al.(2018)Jhamtani, Gangal, Hovy, Neubig, and
  Berg-Kirkpatrick}]{jhamtani-etal-2018-learning}
Harsh Jhamtani, Varun Gangal, Eduard Hovy, Graham Neubig, and Taylor
  Berg-Kirkpatrick. 2018.
\newblock \href {https://doi.org/10.18653/v1/P18-1154} {Learning to generate
  move-by-move commentary for chess games from large-scale social forum data}.
\newblock In \emph{Proceedings of the 56th Annual Meeting of the Association
  for Computational Linguistics (Volume 1: Long Papers)}, pages 1661--1671,
  Melbourne, Australia. Association for Computational Linguistics.

\bibitem[{Kim et~al.(2025)Kim, Goh, Hwang, Cho, and
  Ok}]{kim-etal-2025-bridging}
Jaechang Kim, Jinmin Goh, Inseok Hwang, Jaewoong Cho, and Jungseul Ok. 2025.
\newblock \href {https://doi.org/10.18653/v1/2025.naacl-long.481} {Bridging the
  gap between expert and language models: Concept-guided chess commentary
  generation and evaluation}.
\newblock In \emph{Proceedings of the 2025 Conference of the Nations of the
  Americas Chapter of the Association for Computational Linguistics: Human
  Language Technologies (Volume 1: Long Papers)}, pages 9497--9516,
  Albuquerque, New Mexico. Association for Computational Linguistics.

\bibitem[{Kocsis and Szepesv{\'a}ri(2006)}]{kocsis2006bandit}
Levente Kocsis and Csaba Szepesv{\'a}ri. 2006.
\newblock Bandit based monte-carlo planning.
\newblock In \emph{Proceedings of the 17th European Conference on Machine
  Learning}, ECML'06, pages 282--293, Berlin, Heidelberg. Springer-Verlag.

\bibitem[{Kolasani et~al.(2025)Kolasani, Saplin, Crispino, Montgomery, Davis,
  Zaharia, Wang, and Wang}]{kolasani2025llmchessbenchmarkingreasoning}
Sai Kolasani, Maxim Saplin, Nicholas Crispino, Kyle Montgomery, Jared~Quincy
  Davis, Matei Zaharia, Chi Wang, and Chenguang Wang. 2025.
\newblock \href {http://arxiv.org/abs/2512.01992} {Llm chess: Benchmarking
  reasoning and instruction-following in llms through chess}.

\bibitem[{Krippendorff(1970)}]{krippendorff1970estimating}
Klaus Krippendorff. 1970.
\newblock \href {https://doi.org/10.1177/001316447003000105} {Estimating the
  reliability, systematic error and random error of interval data}.
\newblock \emph{Educational and Psychological Measurement}, 30(1):61--70.

\bibitem[{{Lichess}()}]{lichess}
{Lichess}.
\newblock Lichess.org --- free online chess.
\newblock \url{https://lichess.org}.
\newblock Accessed: 2026-08-01.

\bibitem[{Lin(2004)}]{lin-2004-rouge}
Chin-Yew Lin. 2004.
\newblock \href {https://aclanthology.org/W04-1013/} {{ROUGE}: A package for
  automatic evaluation of summaries}.
\newblock In \emph{Text Summarization Branches Out}, pages 74--81, Barcelona,
  Spain. Association for Computational Linguistics.

\bibitem[{Monroe and Chalmers(2024)}]{monroe2024masteringchesstransformermodel}
Daniel Monroe and Philip~A. Chalmers. 2024.
\newblock \href {http://arxiv.org/abs/2409.12272} {Mastering chess with a
  transformer model}.

\bibitem[{Monroe et~al.(2026)Monroe, Eilender, Chalmers, Tang, and
  Anderson}]{monroe2026chessformerunifiedarchitecturechess}
Daniel Monroe, George Eilender, Philip Chalmers, Zhenwei Tang, and Ashton
  Anderson. 2026.
\newblock \href {http://arxiv.org/abs/2605.19091} {Chessformer: A unified
  architecture for chess modeling}.

\bibitem[{Papineni et~al.(2002)Papineni, Roukos, Ward, and
  Zhu}]{papineni-etal-2002-bleu}
Kishore Papineni, Salim Roukos, Todd Ward, and Wei-Jing Zhu. 2002.
\newblock \href {https://doi.org/10.3115/1073083.1073135} {{B}leu: a method for
  automatic evaluation of machine translation}.
\newblock In \emph{Proceedings of the 40th Annual Meeting of the Association
  for Computational Linguistics}, pages 311--318, Philadelphia, Pennsylvania,
  USA. Association for Computational Linguistics.

\bibitem[{Pascutto and Linscott(2024)}]{lc0}
Gian-Carlo Pascutto and Gary Linscott. 2024.
\newblock Leela chess zero.
\newblock \url{https://lczero.org/}.
\newblock Version 0.31.2.

\bibitem[{Rosin(2011)}]{rosin2011multi_puct}
Christopher~D. Rosin. 2011.
\newblock Multi-armed bandits with episode context.
\newblock \emph{Annals of Mathematics and Artificial Intelligence},
  61(3):203--230.

\bibitem[{Schrittwieser et~al.(2020)Schrittwieser, Antonoglou, Hubert,
  Simonyan, Sifre, Schmitt, Guez, Lockhart, Hassabis, Graepel, Lillicrap, and
  Silver}]{Schrittwieser_2020}
Julian Schrittwieser, Ioannis Antonoglou, Thomas Hubert, Karen Simonyan,
  Laurent Sifre, Simon Schmitt, Arthur Guez, Edward Lockhart, Demis Hassabis,
  Thore Graepel, Timothy Lillicrap, and David Silver. 2020.
\newblock \href {https://doi.org/10.1038/s41586-020-03051-4} {Mastering atari,
  go, chess and shogi by planning with a learned model}.
\newblock \emph{Nature}, 588(7839):604–609.

\bibitem[{Shannon(1950)}]{Shannon1950}
Claude~E. Shannon. 1950.
\newblock Programming a computer for playing chess.
\newblock \emph{Philosophical Magazine}, 41(314):256--275.

\bibitem[{Silver et~al.(2017)Silver, Hubert, Schrittwieser, Antonoglou, Lai,
  Guez, Lanctot, Sifre, Kumaran, Graepel, Lillicrap, Simonyan, and
  Hassabis}]{silver2017masteringchessshogiselfplay}
David Silver, Thomas Hubert, Julian Schrittwieser, Ioannis Antonoglou, Matthew
  Lai, Arthur Guez, Marc Lanctot, Laurent Sifre, Dharshan Kumaran, Thore
  Graepel, Timothy Lillicrap, Karen Simonyan, and Demis Hassabis. 2017.
\newblock \href {http://arxiv.org/abs/1712.01815} {Mastering chess and shogi by
  self-play with a general reinforcement learning algorithm}.

\bibitem[{Tang et~al.(2026)Tang, Wen, Grief-Albert, Elgabra, Yang, Dong, and
  Anderson}]{tang2026groundedchessreasoninglanguage}
Zhenwei Tang, Qianfeng Wen, Seth Grief-Albert, Yahya Elgabra, Blair Yang,
  Honghua Dong, and Ashton Anderson. 2026.
\newblock \href {http://arxiv.org/abs/2603.20510} {Grounded chess reasoning in
  language models via master distillation}.

\bibitem[{{The Stockfish Developers}()}]{stockfish}
{The Stockfish Developers}.
\newblock Stockfish.
\newblock \url{https://github.com/official-stockfish/Stockfish}.
\newblock Open-source chess engine.

\bibitem[{Zhang et~al.(2025{\natexlab{a}})Zhang, Jacob, Lai, Fried, and
  Ippolito}]{zhang2025allie}
Yiming Zhang, Athul~Paul Jacob, Vivian Lai, Daniel Fried, and Daphne Ippolito.
  2025{\natexlab{a}}.
\newblock \href {https://openreview.net/forum?id=bc2H72hGxB} {Human-aligned
  chess with a bit of search}.
\newblock In \emph{The Thirteenth International Conference on Learning
  Representations}.

\bibitem[{Zhang et~al.(2025{\natexlab{b}})Zhang, Han, Li, Chen, and
  Lin}]{zhang-etal-2025-complete}
Yinqi Zhang, Xintian Han, Haolong Li, Kedi Chen, and Shaohui Lin.
  2025{\natexlab{b}}.
\newblock \href {https://doi.org/10.18653/v1/2025.naacl-short.1} {Complete
  chess games enable {LLM} become a chess master}.
\newblock In \emph{Proceedings of the 2025 Conference of the Nations of the
  Americas Chapter of the Association for Computational Linguistics: Human
  Language Technologies (Volume 2: Short Papers)}, pages 1--7, Albuquerque, New
  Mexico. Association for Computational Linguistics.

\end{thebibliography}

\appendix

\section{Case Study -- Qualitative Analysis}
\label{app:casestudy}

\begin{figure}[t]
  \centering
  \includegraphics[width=0.86\columnwidth]{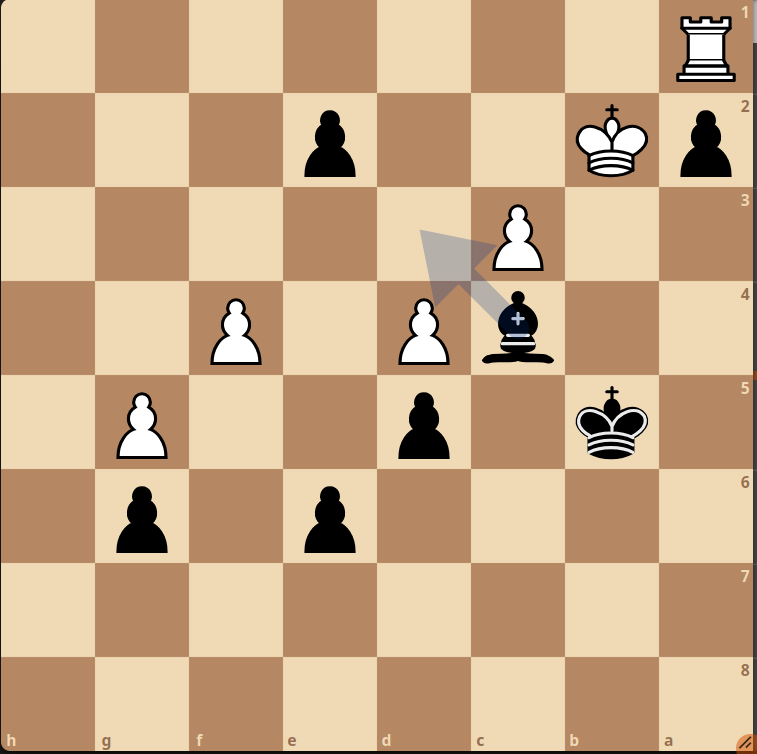}
  \caption{Black to move.
  Lichess puzzle rating: \rating{2540}. The engine's best move is
  \textbf{Bd3}.}
  \label{fig:case-study}
\end{figure}

\noindent\textbf{Position.}
Figure~\ref{fig:case-study} shows a rook-versus-bishop endgame..

\vspace{2pt}
\noindent\textbf{Expert reference explanation.}
A human expert characterises the position as follows: \emph{in order to promote
the pawns, the king has to be brought closer to the pawns, and there is no
obvious way of doing this. Moving the bishop clears the path for the king to
march into the centre of the board and support the promotion of the e2-pawn.
This is the central idea behind} \mbox{Bd3}.

\vspace{2pt}
\noindent\textbf{Setup.}
We compare the recursive strategy and the baseline. Text matching the expert's core idea---\emph{clearing the
bishop's square so the king can approach and escort the e-pawn}---is highlighted
in \hit{green}. The recursive output is abridged with \elide\ for space. Hallucinations are highlighted in \miss{red}.

\begin{outbox}{Recursive strategy (abridged)}
\small
From Black's perspective, the position is dominated by two advanced passed
pawns: the pawn on e2 is one step from promotion, and the pawn on a2 is also
close to queening. The bishop on c4 supports the queenside pawn, while
\hit{Black's king must improve its position so that the pawns can promote
safely.} \elide\ The main problem for Black is therefore not creating a
promotion threat; that already exists. The problem is making promotion safe.
\elide\ \hit{Black must first improve the coordination of the king and bishop,
or create a second threat that forces the rook away.} The strongest move found
by the search is 23\dots Bd3, \elide\ \hit{The immediate purpose of 23\dots Bd3 is to reposition
the bishop so that it supports the e-pawn's promotion plan more effectively.}
\elide\ \hit{Bd3 begins a process in which Black improves the king and bishop
until White is offered a tempting capture that would remove the rook from its
defensive duty.} \elide\ Black's continuation, 24\dots Kc4, shows the point.
\hit{Black brings the king closer to the central pawn and bishop instead of
pushing \dots e1=Q prematurely.} \elide\ Black follows with
\hit{26\dots Kd3, bringing the king still closer}, and after 27.Kc1, Black has
27\dots Bg2. \elide\ Therefore Bd3 is best because it addresses the real issue:
not whether Black can promote, but whether Black can force the rook away from
the promotion square. \hit{It coordinates the bishop with the e-pawn, keeps the
king free to approach}, and creates a future overload or deflection motif.
\end{outbox}

\begin{outbox}{Single-call strategy (abridged)}
\small
Black is to move in a sharp endgame with a bishop, rook, exposed kings, and two
advanced passed pawns on a2 and e2. Black has structural weaknesses, so the
position should be handled dynamically: the passed pawns and active bishop must
compensate for the imperfect pawn formation. The immediate candidates are Bd3,
Ka4, and Kc6. Ka4 and Kc6 are understandable king-improvement moves, but their
continuations allow White's rook to become more active and do not exploit the
bishop's ability to control central squares. Bd3 is filtered out as best because
it increases piece activity, keeps both passed pawns relevant, and forces White
to spend time with Kxa2 before \hit{Black centralizes with Kc4} and continues
Be4. The practical lesson is that in an endgame with advanced passers, activity
and coordination usually matter more than static pawn weaknesses: \hit{improve the
piece that can support promotion} and restrict the enemy king, rather than making
a slow king move.
\end{outbox}

\begin{outbox}{Baseline (vanilla prompting)}
\small
Black is materially ahead in the most important practical sense: the a2- and
e2-pawns are both one step from promotion. However, neither can promote
immediately. The a1-square is occupied by White's rook, while 23\dots e1=Q?
fails to 24.Rxe1 because the rook on a1 controls the entire first rank. Black's
king on b5 and bishop on c4 are active, but the bishop is currently tied to
useful defensive work: it protects the a2-pawn and helps keep the advanced pawns
moving safely. \elide\ A king move improves the king only slowly and does not
solve the central problem that White's rook can attack the promotion squares.
\elide\ 23\dots Bd3! is the useful waiting-and-improving move. The bishop
becomes much more active: from d3 it attacks the \miss{c2-pawn}, controls b1, and
supports the e2-pawn along the d3--e2 diagonal. \elide\ The key motif is
promotion geometry combined with deflection: Black wants to make White move the
rook away from a1 or \miss{force White to spend a tempo defending the c2-pawn.}
\elide\ White should not rush to 24.Rxa2. Although that captures one passed
pawn, it moves the rook off a1, and then 24\dots e1=Q! promotes with no Rxe1
available. \elide\ Compared with a king move, Bd3 creates immediate pressure on
c2 and improves the bishop's activity while preserving Black's winning pawn
threats. \elide\ The important beginner cues are: first, always check whether a
promotion square is controlled before promoting; second, notice that a rook on
a1 is both a defender and a potential target for deflection.
\end{outbox}

\vspace{2pt}
\noindent\textbf{Analysis.}
All systems select the correct move, and all identify the surface-level
tactical constraint that prevents an immediate \dots e1=Q. They diverge on
\emph{how completely} they recover the expert's reasoning. Both tree-grounded
strategies find the expert's core idea---the bishop steps aside so that the
king can approach and escort the e-pawn---but they recover it at different
resolutions. The single-call strategy states the idea in passing: it notes that
Black centralises with \mbox{Kc4} and that the piece supporting promotion
should be improved, but the king march appears as a single consequence of
\mbox{Bd3} rather than as the plan that motivates it. The recursive strategy,
which propagates evaluations bottom-up from the deeper nodes of the subtree,
recovers the plan in full: it identifies king--bishop coordination as the real
problem from the outset, and then traces the concrete march
(24\dots Kc4, 26\dots Kd3) move by move, together with the deflection motif
that makes it work. Because these continuations are grounded in the child nodes
being summarised, the top-level explanation inherits them instead of having to
rediscover them at the root.
The position-only baseline explanation remains at the root: it justifies \mbox{Bd3} in
terms of static piece activity (attacking a pawn on c2, controlling b1) and
explicitly \emph{dismisses} king improvement as too slow, the opposite of the
expert's reading. It also introduces a factual error absent from both
tree-grounded outputs, referring to a \miss{non-existent c2-pawn (the pawn
stands on c3)} and then building its deflection argument on that pawn. This example highlights the limitation of the baseline: without intermediate summarization, strategic intent that only emerges several plies ahead is often missed, which leads the model to generate plausible but hallucinated local justifications.

\section{Additional Experimental Results and Analysis}
\label{app:add-exp-results}

\subsection{Paired Differences}
\label{app:paired-differences}

Table~\ref{tab:paired_diffs} reports paired score differences from the LLM judge, where positive values favor the first strategy. Both SC and Rec significantly outperform all three baselines on correctness, theme, completeness, and contrastiveness, but score lower on pedagogy. SC's advantage is largest on correctness, while Rec's is largest on theme, completeness, and contrastiveness, at the cost of a wider pedagogy gap. Removing Maia-3 at a matched candidate budget mainly reduces contrastiveness and, to a lesser extent, completeness, leaving the other dimensions largely unchanged.

\subsection{What if the LLM Judge is provided the search tree?}
\label{app:judge-with-search-tree}

\providecommand{\ci}[1]{\textcolor{black!45}{#1}}          %
\providecommand{\pd}[3]{$#1$ \ci{[$#2$, $#3$]}}            %

\begin{table*}[t]
\centering
\begin{tabular}{lccc}
\toprule
Dimension & PO & SC & Rec \\
\midrule
chess correctness & \pd{2.76}{2.54}{2.98} & \pd{3.39}{3.21}{3.57} & \pd{\mathbf{3.42}}{3.25}{3.59} \\
theme             & \pd{3.58}{3.33}{3.83} & \pd{4.44}{4.30}{4.57} & \pd{\mathbf{4.61}}{4.46}{4.75} \\
pedagogy          & \pd{3.84}{3.67}{4.01} & \pd{3.76}{3.67}{3.84} & \pd{\mathbf{3.87}}{3.78}{3.96} \\
completeness      & \pd{3.06}{2.79}{3.33} & \pd{3.95}{3.82}{4.07} & \pd{\mathbf{4.42}}{4.24}{4.58} \\
contrastiveness   & \pd{2.86}{2.70}{3.02} & \pd{3.62}{3.50}{3.74} & \pd{\mathbf{3.82}}{3.72}{3.92} \\
\midrule
\textbf{aggregate} & \pd{3.22}{3.03}{3.41} & \pd{3.83}{3.73}{3.93} & \pd{\mathbf{4.03}}{3.92}{4.13} \\
\bottomrule
\end{tabular}
\caption{Mean scores with 95\% confidence intervals in gray. Best per row in bold.}
\label{tab:results-judge-with-tree}
\end{table*}

Table~\ref{tab:results-judge-with-tree} reports the results of the evaluation where the LLM Judge is provided the search tree. We observe that similar trends mostly persist. The correctness of the recursive strategy increases and becomes similar to the single-call strategy.

\subsection{What if puzzles are selected randomly?}
\label{app:random-puzzle-selection}

Table ~\ref{tab:results-random-puzzles} reports the results. The trends are similar to what was observed with popular puzzles.

\providecommand{\ci}[1]{\textcolor{black!45}{#1}}          %
\providecommand{\pd}[3]{$#1$ \ci{[$#2$, $#3$]}}            %

\begin{table*}[t]
\centering
\begin{tabular}{lccc}
\toprule
Dimension & PO & SC & Rec \\
\midrule
chess correctness & \pd{2.63}{2.41}{2.84}          & \pd{\mathbf{3.22}}{3.06}{3.37} & \pd{2.99}{2.84}{3.15} \\
theme             & \pd{3.42}{3.16}{3.68}          & \pd{3.90}{3.73}{4.06}          & \pd{\mathbf{4.35}}{4.18}{4.52} \\
pedagogy          & \pd{\mathbf{3.74}}{3.57}{3.91} & \pd{3.41}{3.30}{3.52}          & \pd{3.50}{3.39}{3.61} \\
completeness      & \pd{2.90}{2.63}{3.17}          & \pd{3.37}{3.19}{3.55}          & \pd{\mathbf{4.04}}{3.82}{4.25} \\
contrastiveness   & \pd{3.07}{2.84}{3.30}          & \pd{3.26}{3.13}{3.39}          & \pd{\mathbf{3.76}}{3.62}{3.90} \\
\midrule
\textbf{aggregate} & \pd{3.15}{2.95}{3.35}         & \pd{3.43}{3.32}{3.55}          & \pd{\mathbf{3.73}}{3.59}{3.86} \\
\bottomrule
\end{tabular}
\caption{Mean scores on randomly sampled puzzles (rather than the most popular ones), with 95\% confidence intervals in gray. Best per row in bold.}
\label{tab:results-random-puzzles}
\end{table*}

\subsection{Cost Analysis}
\label{app:cost-analysis}
To compare the computational cost of the single-call and recursive strategies, we measure their total token consumption. Across all 100 evaluation puzzles, the single-call strategy consumes approximately $0.8$M tokens with GPT-5.6 Luna, whereas the recursive strategy requires approximately $5.4$M tokens. Thus, the single-call strategy reduces token usage by nearly $85\%$, making it substantially more token-efficient. This efficiency, however, comes at the cost of lower performance than the recursive strategy.

\subsection{Qualitative Analysis of Human and LLM-Judge Disagreements}
\label{app:human-llm-disagreement-analysis}

Before sharing the qualitative analysis, we would like the readers to note that identifying a move's underlying motif is inherently subjective, as evidenced by disagreements among human evaluators themselves. Thus, human--LLM disagreements do not necessarily indicate unsupported LLM judgments. We share some of our findings below.

\begin{itemize}
    \item \textbf{Pedagogy Score Deflation:} Search-tree-grounded explanations occasionally reference internal search data (e.g., Q-values, visit counts, ``policy preference''). Human raters tend to overlook such technical terms when scoring, whereas the LLM judge penalizes them consistently.

    \item \textbf{LLM Judge Strictness:} The LLM judge occasionally penalizes explanations that human evaluators would consider adequate. For instance, when a king has no legal moves due to opponent coverage, stating ``no legal moves exist’’ was marked down for omitting move-by-move justification.

    \item \textbf{LLM Judges Catch Peripheral Errors:} When evaluating long-form explanations, human annotators tend to overlook minor errors as long as the central idea is conveyed correctly. LLM judges appear less susceptible to this effect and more often flag errors in details outside the main line of reasoning, such as a slight error listed in a secondary variation.
\end{itemize}

\providecommand{\ci}[1]{\textcolor{black!45}{#1}}          %
\providecommand{\pd}[3]{$#1$ \ci{[$#2$, $#3$]}}                %

\providecommand{\ci}[1]{\textcolor{black!45}{#1}}          %
\providecommand{\pd}[3]{$#1$ \ci{[$#2$, $#3$]}}                %

\begin{table*}[t]
\centering
\scriptsize
\setlength{\tabcolsep}{2pt}
\renewcommand{\arraystretch}{0.92}
\begin{tabular}{l ccccc}
\toprule
\textbf{Comparison} & \textbf{Correctness} & \textbf{Theme} & \textbf{Pedagogy} & \textbf{Completeness} & \textbf{Contrastiveness} \\
\midrule
SC vs.\ PO   & \pd{+0.60}{+0.35}{+0.85} & \pd{+0.57}{+0.33}{+0.81} & \pd{-0.25}{-0.43}{-0.07} & \pd{+0.41}{+0.11}{+0.71} & \pd{+0.22}{+0.10}{+0.32} \\
SC vs.\ PV   & \pd{+0.49}{+0.24}{+0.74} & \pd{+0.17}{+0.05}{+0.29} & \pd{-0.46}{-0.62}{-0.30} & \pd{+0.20}{+0.03}{+0.37} & \pd{+0.20}{+0.05}{+0.35} \\
SC vs.\ CCC  & \pd{+0.52}{+0.26}{+0.78} & \pd{+0.49}{+0.26}{+0.72} & \pd{-0.32}{-0.50}{-0.14} & \pd{+0.40}{+0.12}{+0.68} & \pd{+0.87}{+0.63}{+1.11} \\
SC vs.\ SC w/o Maia-3$^\ast$ & \pd{+0.00}{-0.19}{+0.20} & \pd{+0.02}{-0.15}{+0.20} & \pd{+0.07}{-0.07}{+0.22} & \pd{+0.19}{+0.01}{+0.38} & \pd{+0.14}{+0.04}{+0.24} \\
\midrule
Rec vs.\ PO  & \pd{+0.21}{+0.07}{+0.35} & \pd{+0.82}{+0.56}{+1.08} & \pd{-0.56}{-0.73}{-0.39} & \pd{+0.93}{+0.63}{+1.23} & \pd{+0.59}{+0.35}{+0.83} \\
Rec vs.\ PV  & \pd{+0.10}{+0.02}{+0.18} & \pd{+0.42}{+0.22}{+0.62} & \pd{-0.77}{-0.91}{-0.63} & \pd{+0.72}{+0.35}{+1.09} & \pd{+0.57}{+0.32}{+0.82} \\
Rec vs.\ CCC & \pd{+0.13}{+0.05}{+0.21} & \pd{+0.74}{+0.49}{+0.99} & \pd{-0.63}{-0.79}{-0.47} & \pd{+0.92}{+0.61}{+1.23} & \pd{+1.24}{+1.02}{+1.46} \\
Rec vs.\ Rec w/o Maia-3$^\ast$ & \pd{+0.05}{-0.19}{+0.29} & \pd{+0.03}{-0.16}{+0.22} & \pd{+0.08}{-0.08}{+0.24} & \pd{+0.22}{+0.00}{+0.43} & \pd{+0.36}{+0.19}{+0.53} \\
\bottomrule
\end{tabular}
\caption{Paired score differences (GPT-5.6 Luna, LLM judge) with 95\% CIs in gray; positive values favor the first strategy. All comparisons against the baselines (PO, PV, CCC) are significant at $p<0.05$, whereas the Maia-3 ablations are significant only on a subset of dimensions. $^\ast$Matched candidate budget.}
\label{tab:paired_diffs}
\end{table*}

\section{Model Selection}
\label{app:modelselection}

Our initial experiments used GPT-5.5-High as the automatic judge, on the assumption that
a stronger reasoning model would yield more reliable scores. Preliminary experiments
made this option impractical. Scoring a single explanation consumed roughly
$100K$ tokens, the bulk of which was attributable to the model's internal
reasoning rather than to the prompt or the emitted rubric scores. At a list price of
approximately \$40 per million tokens, this amounts to about \$4 per judgment, so
evaluating our full set of explanations, together with the repeated runs needed to
estimate scoring variance, would have cost more than the project could absorb.
We therefore turned to GPT-5.5-Low because it was financially tractable. Following a similar analysis, we pick GPT-5.6-Luna as the underlying explanation generation model to run most of the experiments.

\section{Evaluation}
\label{app:eval}

\subsection{Evaluation Dimensions}
\label{app:evaldimensions}
\begin{itemize}
    \item \textbf{Chess Correctness}: Whether the explanation accurately describes the position, move consequences, and tactical or strategic claims. This dimension measures if the explanation is consistent with the actual chess position and search evidence.

    \item \textbf{Chess Theme}: Whether the explanation identifies the relevant chess concepts behind the move, such as tactical motifs, king safety, material considerations, piece activity, pawn structure, or long-term strategic ideas.

    \item \textbf{Pedagogy}: Whether the explanation is useful for learning, clearly communicates the underlying idea, and helps a player understand why the move is strong.

    \item \textbf{Completeness}: Whether the explanation covers the key reasons behind the move, including important tactical ideas, strategic motivations, and consequences when necessary.

    \item \textbf{Contrastiveness}: Whether the explanation provides meaningful comparisons with alternative moves and explains why plausible alternatives are inferior.
\end{itemize}

\subsection{Human Evaluation Protocol}
\label{app:humanevalprotocol}

Human raters scored explanations in a custom-built annotation interface. Each
item presented the puzzle position as a board image alongside a link to the
corresponding Lichess analysis board, so that raters could verify claims by
playing through lines themselves rather than relying on the given solution.
Raters scored the five dimensions mentioned in Appendix~\ref{app:evaldimensions}.

\begin{promptbox}[title={Instructions shown to human raters}]
\begin{prompt}
For each item you'll see a puzzle position, the best move, and an explanation of
why it's best. Use the board image and the Lichess analysis link to check the
explanation's claims for yourself, then score it against each of the five criteria
below.

\textbf{Judge every claim against the position on the board, not against how
closely it follows the given solution line---an explanation that reaches the same
conclusion by a different, correct route is not wrong.}

All five scores are required before you can save and move to the next item.
\end{prompt}
\end{promptbox}

\begin{promptbox}[title={Rater-facing rubric: criteria and score-band anchors}]
\begin{lstlisting}[style=promptstyle]
Chess correctness -- Are the chess facts and claims in the explanation actually
true of this position?
  5 -- Every claim is accurate.
  4 -- All major claims true; one minor imprecision.
  3 -- One non-trivial claim is wrong, but the main idea still holds.
  2 -- Several small inaccuracies, or one that would mislead a reader.
  1 -- Misdescribes the position, move, or tactic.

Theme -- Does the explanation correctly identify and center the chess concept
driving this position?
  5 -- Centers the right concept, applied specifically to this position.
  4 -- Right concept, but a bit generic or with a minor gap.
  3 -- Only partially addresses the theme.
  2 -- Centers a different, tangential concept.
  1 -- Never engages with the right concept, or asserts a wrong one.

Pedagogy -- Is the explanation well-suited to teaching a {player_level} -- right level,
no unexplained jargon?
  5 -- Builds from what's visible on the board to the insight; jargon-free or
       glossed.
  4 -- Mostly {player_level}-appropriate, one small friction point.
  3 -- Uneven -- some parts well-scaffolded, others assume prior knowledge.
  2 -- Poor scaffolding -- multiple unglossed terms or unexplained conclusions.
  1 -- Dense engine/notation-speak; a {player_level} would learn nothing usable.

Completeness -- Is the core reason the move is best actually covered?
  5 -- The central reason is clearly and completely stated.
  4 -- Core reason identified; one secondary supporting point missing.
  3 -- Touches the right idea but stops short of completing it.
  2 -- The right idea is only glancingly mentioned.
  1 -- Never articulates the actual reason this move is best.

Contrastiveness -- Does the explanation address why this move beats the
alternatives a player would consider?
  5 -- Names a genuinely tempting alternative and explains concretely why it
       falls short.
  4 -- Addresses a tempting alternative with basically right, if generic,
       reasoning.
  3 -- Names an alternative but doesn't really engage with why it's worse.
  2 -- Another option is only acknowledged in passing.
  1 -- Describes the chosen move in isolation, no engagement with alternatives.
\end{lstlisting}
\end{promptbox}

\section{Prompts}
\label{app:prompts}

\subsection{LLM-as-a-Judge Prompt}
\label{app:judge-prompt}

\begin{promptbox}[title={LLM-as-a-judge prompt}]
\begin{lstlisting}[style=promptstyle]
You are an expert chess coach and evaluator (FIDE-master strength). Score the
explanation below on how well it teaches why the given move is best.

Position (FEN): {fen}
Best move: {best_move_san}

Solution line (the puzzle's own solution, starting with the move above):
{solution_line_san}
Reference theme/concept label: {themes}

Explanation to judge:
"""
{reasoning}
"""

How to judge:
Before reading the explanation, set up the position from the FEN yourself and do your own analysis of it:
1. Work out why {best_move_san} is best -- the point the solution line above turns on. If your analysis says some other move is just as good, note it: that makes it a fair candidate for the explanation to have discussed, not a mistake in the explanation.
2. Work out the plausible alternatives to {best_move_san}: the moves a player at the board would genuinely be tempted to play instead. Include the natural human tries -- an obvious check, capture, recapture, developing or defensive move that looks right and falls short -- not only the second-best engine move. For each one, work out concretely why it is worse than {best_move_san}: the refutation, the line, or what it concedes. You are given no engine output, search statistics, or alternative-move evidence: this list of alternatives and their refutations is yours to derive, and it is the
reference you score contrastiveness against. Do not take the explanation's word for any of it -- if the explanation names an alternative, verify on the board that the move is legal, that it is genuinely tempting, and that the stated reason it fails actually holds.

The move, the solution line, and the theme label above are a strong reference, not an absolute standard. They come from a puzzle database: the line can stop early or run through a defence that is not the toughest, another move can be just as good, and the theme label can be coarse or miss an idea that is just as present in the position. Treat them as the starting point of your own analysis rather than an answer key.

So judge the explanation on its merit as an explanation of this position, not on how closely it echoes the reference: is what it claims true on this board, does it capture the idea that actually makes the move work, does it engage with what a player would really be tempted to play instead? An explanation that departs from
the reference wording and is right about the position keeps full credit; one that matches the reference while getting the position wrong does not earn it. Fluency, confidence, and plausible-sounding phrasing are not evidence of any of this -- check the claims.

Then score the explanation on every criterion below from 1 (very poor) to 5 (excellent). Score the criteria strictly in the order they are listed: judging chess_correctness first anchors you in the object-level chess facts before the softer, more subjective dimensions that follow. For every criterion, write your reasoning before you decide the score, and ground it in concrete evidence from the explanation and, for chess_correctness and contrastiveness, in a specific move -- a solution-line move, or a named alternative from your own analysis -- rather than a generic statement.

Criteria:
{criteria}

Output JSON with exactly one key, "scores": a list with exactly one entry per
criterion, in the order given, each entry an object with exactly these keys, in
this order:
- criterion (one of: {criterion_names})
- reasoning (2-4 sentences, written before the score, citing concrete evidence)
- score (integer 1-5)
\end{lstlisting}
\end{promptbox}

\subsection{Leaf-Node Explanation Prompt}
\label{app:leaf-prompt}

\begin{promptbox}[title={Leaf-node prompt (bottom-up strategy, \texttt{none} grounding)}]
\begin{lstlisting}[style=promptstyle]
You are a strong chess coach explaining a single move to a serious club player.

Position before the move (FEN): {parent_fen}
Resulting position after the move (FEN): {fen}
Move sequence from the root: {moves_from_root}
The engine's top move here: {move_san}, played by {mover}
Search value estimate for this move (from {mover}'s perspective): {q}
Other candidate moves that suggest themselves: {sibling_moves}

Task:
Explain, from {mover}'s perspective, why {move_san} is best in a way a serious
club player could find over the board. First name the key strategic and tactical
features of the position and the candidate moves that naturally suggest
themselves. Then discuss the thought process leading to {move_san} -- the
pattern, motif, or imbalance that makes it work. Compare it with the most
plausible alternatives and say precisely why they are inferior. Emphasize the
practical cues a player should notice so the explanation builds intuition rather
than merely reporting engine variations. Include concrete variations only
insofar as they clarify the ideas.

Rules:
1. Use only the evidence given above.
2. Do not assert a concrete tactic (a fork, a pin, a forced mate, winning
   material) unless you can verify it directly from the position given.
3. Write from {mover}'s perspective -- this is {mover}'s move.

Output JSON with exactly one key:
- reasoning (a single prose string)
\end{lstlisting}
\end{promptbox}

\subsection{Internal-Node Aggregation Prompt}
\label{app:internal-prompt}

\begin{promptbox}[title={Internal-node aggregation prompt (bottom-up strategy)}]
\begin{lstlisting}[style=promptstyle]
You are a strong chess coach and reasoning synthesizer grounded in an MCTS
search tree.

Position after the move (FEN): {fen}
Move that reached this position: {move_san}, played by {mover}
Learner level: {learner_level}
Continuation from here (principal variation): {pv_san}

The opponent ({opponent}) has just been reached; the search explored their
replies below.
Opponent's strongest reply: {best_reply_san}
  N={best_n}, Q={best_q}, P={best_p}
  Its reasoning (from {opponent}'s perspective): "{best_reply_reasoning}"

Alternative candidates considered:
{alternatives_block}

Task:
Start by clearly describing the key chess properties of the resulting position
from {mover}'s perspective -- what {move_san} was trying to achieve. Then cover
{opponent}'s immediately plausible replies below: {best_reply_san} and the
alternatives above. Using each reply's own reasoning, analyze what it achieves
for {opponent}. Explicitly discuss why every reply, including the strongest one
({best_reply_san}), fails to refute {move_san} -- this is the point of reasoning
bottom-up: a move that looked quiet may turn out to win material or force a
favorable transition several plies later. Then explain why {move_san} is
therefore a strong move: its immediate purpose, the pattern or motif that makes
it effective, and the deeper causal storyline through which {mover}'s advantage
materializes. Preserve concrete findings from the replies' reasoning (e.g. a
capture, a check, a mating net, a structural gain, etc.) rather than letting
them vanish into generalities. Use only the evidence given -- the positions, the
search statistics, and any feature hints above. However, your response shouldn't
mention them. Your response should rely on human understandable chess concepts.
Finish with the practical cues a {learner_level}-level player should notice so
the explanation builds transferable intuition.

Rules:
1. Use only the evidence given -- the children's reasoning and the statistics
   above.
2. Perspective: the reasoning handed to you for each reply below is written from
   {opponent}'s side; you are explaining {move_san} from {mover}'s side. Restate
   everything from {mover}'s perspective.
3. Base every claim only on the supplied position and search tree; state
   uncertainty explicitly when the evidence is shallow or contradictory.
4. Keep the explanation to about {reasoning_word_budget} words.

Output JSON with exactly one key:
- reasoning (a single prose string)
\end{lstlisting}
\end{promptbox}

\subsection{Single-Call Prompt}
\label{app:single-call-prompt}

\begin{promptbox}[title={Single-call strategy prompt}]
\begin{lstlisting}[style=promptstyle]
You are a strong chess coach and reasoning synthesizer grounded in an MCTS
search tree.

You are given the engine's pruned search tree below. {tree_shape_sentence} Every
node lists the move that reaches it, who played it, and the search statistics N
(visits), Q (value from that mover's perspective), and P (policy prior); some
nodes also carry mechanically detected feature hints.

Root position -- {side} to move. You are choosing {side}'s move here; the
best-marked first candidate is the engine's recommendation.
{human_candidates_note}

Search tree:
{serialized_tree}

Task:
Reason about the tree from the leaves up to the root, exactly as a bottom-up
analysis would:
1. For each frontier node (a leaf, with no replies below it), explain -- from
   the perspective of the side that is about to play -- why it is strong or weak
   using chess concepts.
2. For each node that has children, start by clearly describing the key chess
   properties of the board position. Then cover what the immediately plausible
   moves are (children of this node). Afterwards, use the reasoning of each
   children to analyze what each of those moves achieve. Explicitly discuss the
   issues with all the alternative moves. Then explain how the best move can be
   filtered out. How can it be explained from the board position using chess
   concepts? The merged explanation shouldn't lose track of the longer story
   generated by the children.
3. Finally, at the root, start by clearly describing the key chess properties of
   the board position. Then cover what the immediately plausible moves are
   (children of this node). Afterwards, use the reasoning of each children to
   analyze what each of those moves achieve. Explicitly discuss the issues with
   all the alternative moves. Then explain how the best move can be filtered
   out. At every step, if possible, use chess concepts to strengthen the
   intuition behind the analysis. Finish with the practical cues a
   {learner_level}-level player should notice so the explanation builds
   transferable intuition.

Perspective: a node's children are the opponent's replies to it, so their
reasoning is written from the opponent's side; restate everything from the
mover's side as you fold replies upward. At the root the candidates are {side}'s
own moves, so no flip is needed there.

Rules:
1. Use only the evidence given -- the positions, the search statistics, and any
   feature hints above. However, your response shouldn't mention them. Your
   response should rely on human understandable chess concepts.
2. Do not assert a concrete tactic (a fork, a pin, a forced mate, winning
   material) you cannot justify from that evidence; where feature hints are
   given, do not contradict them.
3. Use the MCTS visits and values to resolve conflicting evidence, but do not
   merely report statistics or engine lines.

Output JSON with exactly two keys:
- steps: a list of objects, each with:
    - move: the node's move in SAN (or "root" for the root's synthesis)
    - reasoning: the reasoning for that move, from its mover's perspective
  Order the list leaves first and the root last, reflecting the bottom-up
  procedure. For each node, the reasoning should be detailed, start by clearly
  describing the key chess properties of the board position. Then cover what the
  immediately plausible moves are (children of this node in the search tree).
  Afterwards, use the reasoning of each children to analyze what each of those
  moves achieve. Explicitly discuss the issues with all the alternative moves.
  Then explain how the best move can be filtered out. How can it be explained
  from the board position using chess concepts? The merged explanation shouldn't
  lose track of the longer story generated by the children. At least 5 sentences
  per node should be used.
- root_reasoning: the final root explanation from step 3.
\end{lstlisting}
\end{promptbox}

\subsection{Claim-Level Verification Prompt}
\label{app:claim-level-verification-prompt}

\begin{promptbox}[title={Claim extraction prompt}]
\begin{lstlisting}[style=promptstyle]
You convert a chess explanation into atomic claims that a rule engine will check.
You transcribe; you never judge. Do not decide whether a claim is true, and never
correct a square, piece or move from the explanation, even if it looks wrong.

Root position (FEN): {root_fen}
Side to move: {side_to_move}
Board (uppercase = White, lowercase = Black):
{ascii_board}
Move being explained: {best_move}

Explanation:
"""
{explanation}
"""

Tools (use only these; argument names and types must match exactly):
{tool_catalog}

Instructions
1. Find every statement the tools above check exactly (see 13). Skip plans,
   evaluations, intentions, "best" or "strong", and anything about moves before
   the root.
   Put the number of skipped statements in skipped_unverifiable.
2. Make one atom per assertion. Split compound sentences.
3. For each atom, write `line`: the SAN moves from the root to the position the
   assertion is about. Use earlier sentences to recover implied moves. Ignore
   move numbers in the text. Hypothetical lines ("if X, then Y") are lines.
4. Decompose each atom into sub-claims. Each sub-claim is one tool call with
   `expected` true or false. Put moves that apply to one sub-claim only in its
   `continuation`; use "--" for a pass when expressing a threat.
5. Every piece the text names or implies on a square becomes a piece_on sub-claim.
6. Negated statements use expected: false. Colours are always white/black.
7. Write SAN without move numbers, annotations, "+" or "#".
8. Copy source_span word for word from the explanation. The FEN, side to move
   and board above are context, not claims: never make an atom from them.
9. Transcribe, don't judge: extract what the text asserts even if it looks wrong.
   Use the board only to resolve references ("the rook" -> the square it stands
   on) and to write SAN; never move a claim to a different square or piece.
10. "Forced" maps to is_only_legal_move only when the text says it is the only
    move; otherwise skip it. A claim repeated in the text is extracted once.
11. Number atoms a1, a2, ... and sub-claims a1.s1, a1.s2, ...
12. Each sub-claim is checked in the position reached by playing `line`, then
    its `continuation`, from the root. A `move` argument is a move by the side
    to move in THAT position; never repeat a move that is already in `line`.
    In the example: "d5 is legal" is line [] with is_legal(d5); "after d5 the
    pawn attacks c6" is line ["d5"] with attacks(d5, c6).
13. Make an atom only when its sub-claims check exactly what the statement
    says. If the closest tool checks something different, weaker or stronger,
    give the statement no atom and count it in skipped_unverifiable. Skipping
    is always fine; a check that does not match its statement is a mistake.
    For example, "X fails" or "X does not work" is not "X is illegal", listing
    X among the options does not say X is legal, and "cannot capture the pawn"
    is not "nothing attacks the pawn". The phrase guide's mappings count as
    exact. Never pad an atom with facts the text does not state.
14. Copy source_span character for character, including capital letters.
15. The example is about a different position. Never copy its claims; extract
    only what the explanation above says.
16. Check the board for each piece's colour: uppercase letters are White's
    pieces, lowercase are Black's.

Phrase guide:
{phrase_guide}

Example:
{worked_example}

Return only JSON matching this schema:
{json_schema}
\end{lstlisting}
\end{promptbox}

\end{document}